\documentclass[runningheads]{llncs}
\usepackage{eccv}

\usepackage{eccvabbrv}

\usepackage{graphicx}
\usepackage{booktabs}
\usepackage{soul}
\usepackage{colortbl}
\usepackage{makecell}
\usepackage{multirow} 
\usepackage{wrapfig}

\usepackage{capt-of}

\usepackage[accsupp]{axessibility}  

\usepackage{hyperref}

\usepackage{orcidlink}

\begin{document}

\title{DISPLAY: Directable Human-Object Interaction Video Generation via Sparse Motion Guidance and Multi-Task Auxiliary} 

\titlerunning{DISPLAY}

\author{
Jiazhi Guan$^*$ \and
Quanwei Yang$^*$ \and
Luying Huang \and
Junhao Liang \and
Borong Liang \and
Haocheng Feng \and
Wei He \and
Kaisiyuan Wang$^{\dag}$ \and
Hang Zhou$^{\dag}$ \and
Jingdong Wang
}

\authorrunning{J. Guan et al.}

\institute{
Baidu Inc. \\
\email{\{guanjiazhi,wangkaisiyuan,zhouhang09\}@baidu.com}
}

\maketitle
\def\thefootnote{*}\footnotetext{Equal Contribution.}
\def\thefootnote{\dag}\footnotetext{Corresponding Authors.}
\def\thefootnote{\arabic{footnote}}

\begin{figure}[h!]
\includegraphics[width=1.0\linewidth]{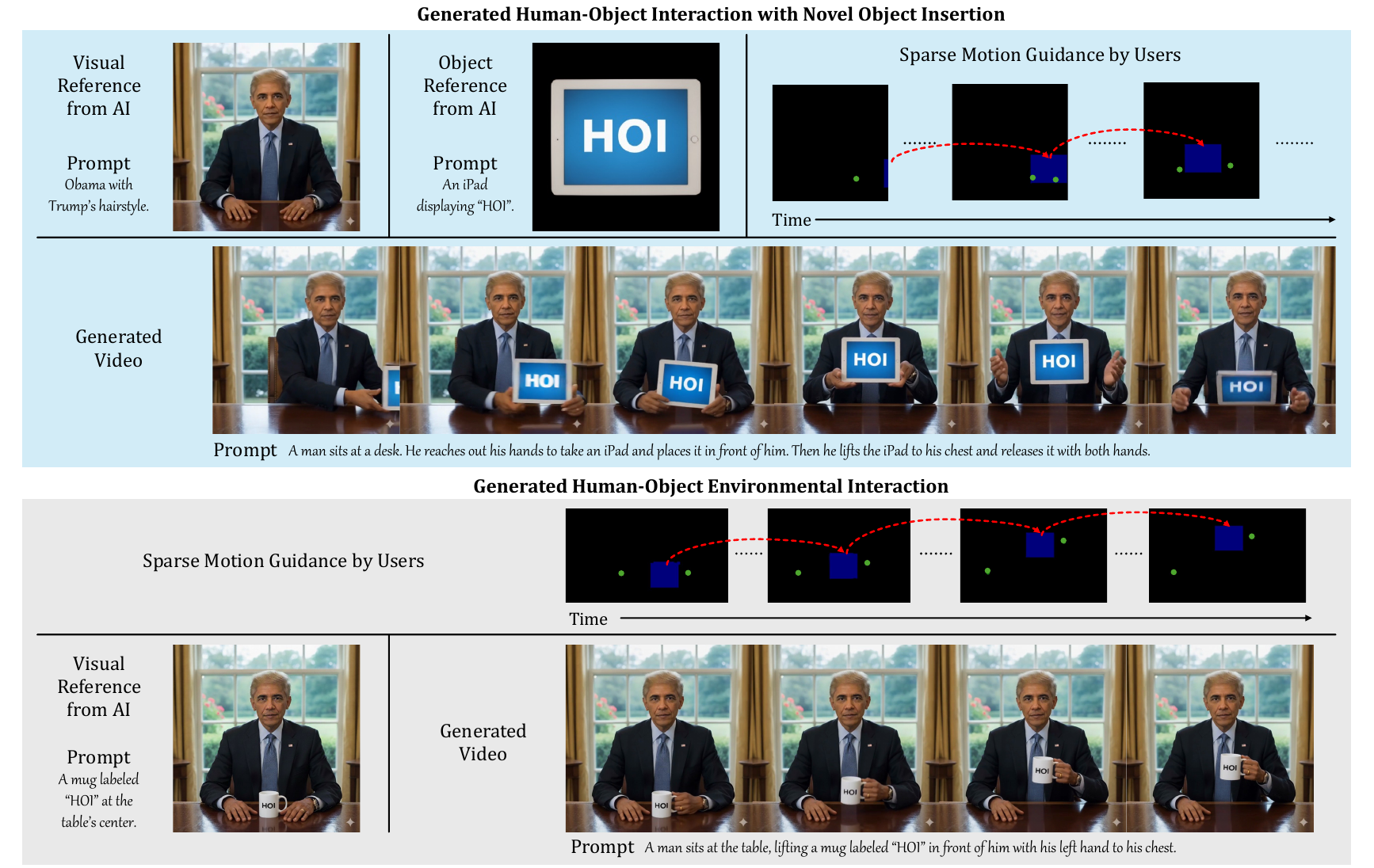}
\captionof{figure}{
\textbf{Human-Object Interactions Synthesis}. 
Our method generates user-intended human-object interactions from zero-shot references and user-specified sparse motion guidance.
Blue: a generated video performing human-object interactions with novel visual and object references.
Gray: a generated video demonstrating environmental interactions conditioned solely on the visual reference.
}
 \label{fig:teaser}
\end{figure}

\begin{abstract}
Human-centric video generation has advanced rapidly, yet existing methods struggle to produce controllable and physically consistent Human-Object Interaction (HOI) videos. Existing works rely on dense control signals, template videos, or carefully crafted text prompts, which limit flexibility and generalization to novel objects.
We introduce a framework, namely DISPLAY, guided by \textit{Sparse Motion Guidance}, composed only of wrist joint coordinates and a shape-agnostic object bounding box. This lightweight guidance alleviates the imbalance between human and object representations and enables intuitive user control. To enhance fidelity under such sparse conditions, we propose an \textit{Object-Stressed Attention} mechanism that improves object robustness.
To address the scarcity of high-quality HOI data, we further develop a \textit{Multi-Task Auxiliary Training} strategy with a dedicated data curation pipeline, allowing the model to benefit from both reliable HOI samples and auxiliary tasks.
Comprehensive experiments show that our method achieves high-fidelity, controllable HOI generation across diverse tasks.
The project page can be found at \href{https://mumuwei.github.io/DISPLAY/}{\textcolor{blue}{https://mumuwei.github.io/DISPLAY/}}.
\end{abstract}    
\section{Introduction}
\label{sec:intro}
Human-centric video generation is becoming pervasive in our daily lives, empowering a diverse array of service scenarios such as news media, e-commerce, education, and multimedia entertainment. This expansion has been paralleled by a substantial surge in research focusing on the synthesis of human faces~\cite{tian2024emo,vpgc,guan2023stylesync,guan2024resyncer,xu2024hallo} and human bodies~\cite{guan2024talk,huang2024make,yang2024showmaker,guan2025audcast,hu2023animate,echomimicv2} in the past few years.
However, the majority of existing human video generation approaches are limited to simple actions like lip-synchronization or basic gestures. To significantly enhance user experience, there is a growing need to embed more interactive functionalities in digital humans, particularly the ability to create videos
with controllable human-object interaction.

Recently, large video generation models (LVGM)~\cite{blattmann2023stable,yang2024cogvideox,wanvideo,hunyuanvideo} have allowed users to generate highly creative content conditional on specific text prompts. Despite their powerful capabilities, these large models still exhibit a fundamental over-reliance on meticulously crafted text prompts, and the generation process is often non-deterministic. 
While several variants~\cite{vace,kaleido,opens2v,phantom} have explored integrating multiple concepts, their outputs frequently suffer from physical inconsistencies in interactions and generally lack precise control.
Moreover, achieving fine-grained and spatially precise control—such as directing a subject to grasp a specific object at an exact location—is exceptionally challenging when relying solely on textual inputs.
This inherent deficiency in spatio-temporal controllability prevents users from accurately dictating complex Human-Object Interaction (HOI) tasks, which severely restricts the practical deployment and application scope of these models.

On the other hand, several studies~\cite{rehold,anchorcrafter,homa,hoi-swap,vace} have taken initial steps toward achieving more controllable HOI synthesis through explicit guidance.
These methods typically operate within a video-to-video inpainting framework~\cite{rehold,hoi-swap} or a pose-guided animation setting~\cite{anchorcrafter,homa}, which require complex and high-dimensional control signals.
Nevertheless, these approaches face two critical limitations.
First, a key imbalance exists in the representation of the interacting components: while these methods frequently employ strong control signals for the human hand regions (e.g., 2D human pose keypoints, 3D hand meshes), the interacting objects typically lack similarly explicit structural representations.
This representation asymmetry causes the generation models to frequently overfit to the dominant control signals, leading to failure modes such as geometric interpenetration and object deformation, particularly on novel or unseen items.
Second, the necessity of a template video or HOI representations (e.g., human poses~\cite{homa,rehold} and object depth maps~\cite{anchorcrafter}) extracted from a driving source imposes significant constraints on these approaches. This dependency restricts editing capabilities to mere modifications of existing footage, thus preventing the freedom to arbitrarily generate content according to higher-level user demands.

In this paper, we propose a \textbf{D}irectable human-object \textbf{I}nteraction video generation framework via \textbf{SP}arse motion guidance and mu\textbf{L}ti-task \textbf{A}uxiliar\textbf{Y}, named \textbf{DISPLAY}, which tackles the aforementioned problems.
Our key insight is to introduce \textit{Sparse Motion Guidance} that enables more robust and high-fidelity HOI generation with elevated degrees of freedom.
Concretely, the \textit{Sparse Motion Guidance} is composed of the coordinates of hand wrist joints and a shape-agnostic bounding box of the target object, which offers two significant advantages:
1) The sparse guidance effectively alleviates the imbalance between hand and object representations, as well as the discrepancy between training and inference. 
On one hand, neither the wrist points nor the shape-agnostic boxes dominate the training, avoiding over-dense priors on hand gestures or object boundaries. 
On the other hand, this form of guidance ensures consistent representations during inference, even when novel objects with significantly different shapes are introduced.
2) In contrast to methods~\cite{homa,anchorcrafter,rehold} that depend on template or driving videos, our sparse guidance requires only minimal user input, such as simple clicking on a canvas at a few key frames. This lightweight interaction enables intuitive generation without relying on external video sources or complex representations, thereby offering substantially greater flexibility.

However, directly adopting such a sparse condition, while beneficial for control freedom, introduces two interlinked challenges:
1) The inherent sparsity of the guidance inevitably raises the difficulty of learning fine-grained generation, particularly within the complex, localized HOI regions.
2) HOI data are frequently compromised by issues such as occlusion. Consequently, the amount of clean and usable HOI data remains limited, constraining the model’s generalization capability.
To address the first issue, we propose the \textit{Object-Stressed Attention} mechanism, which applies weighted coefficients to emphasize object reference tokens and their interactions during HOI modeling, thereby enhancing the quality of HOI generation.
In terms of the data quantity issue, we propose a novel \textit{Multi-Task Auxiliary Training} strategy together with a carefully designed data curation pipeline. 
By leveraging this mixed training paradigm, the model learns more effectively from a broader and more diverse data corpus, which in turn enhances all inference modalities, including object replacement, insertion, and environmental interaction.

Our contributions can be summarized as follows:
\textbf{1)} We propose a novel framework anchored by \textit{Sparse Motion Guidance} that enables arbitrary, high-fidelity, and robust Human-Object Interaction (HOI) generation, supporting effortless user interaction.
\textbf{2)} We introduce the \textit{Object-Stressed Attention} mechanism to enhance object synthesis robustness under sparse guidance, which guarantees the generated objects are physically consistent with the surrounding scene and human pose.
\textbf{3)} We propose a comprehensive Multi-Task Auxiliary Training strategy, paired with a corresponding data curation pipeline, specifically to overcome the bottleneck imposed by the scarcity of high-quality HOI data.
\section{Related Works}
\label{sec:related}

\noindent\textbf{Human Video Animation}.
Human video animation generation has become a widely studied research topic.
Early GAN-based works~\cite{body1,body2,body3,body3d,body4,body5,body6,fomm,liquid} achieve the fusion of appearance and driving pose information through warping and attention operations. 
Recent years, the remarkable generative capacity of diffusion models has elevated this task to an entirely new level.
A series of works~\cite{hu2023animate,xu2023magicanimate,chang2023magicpose,karras2023dreampose,Champ,tu2025stableanimator}
integrate appearance details and driving pose information within a denoising U-Net to achieve vivid reenactments.
More recently, with the significant advancement of video generation~\cite{wanvideo,yang2024cogvideox,hunyuanvideo}, DiT-based human animation models~\cite{Wan-Animate,RealisDance-DiT,DreamActor-M1,HumanDiT} have greatly improved the realism and temporal coherence of generated characters.
However, these models primarily focus on maintaining overall appearance fidelity and temporal consistency, while overlooking the generation of human–object interactions.

\noindent\textbf{Human-Object Interaction}.
With advances in human-centric generation, the more challenging task of Human–Object Interaction (HOI) generation is drawing growing research attention.
Some studies~\cite{EasyHOI,DiffHOI,HOIAnimator,CG-HOI,InterDiff,GraspXL} focus on 3D reconstruction and generation within HOI.
EasyHOI~\cite{EasyHOI} explored reconstructing HOI from single-view RGB images using large models.
For HOI image/video generation~\cite{HOIimage,hoi-swap,homa,anchorcrafter,Affordance,rehold,VirtualModel,ManiVideo,DreamActor-H1}, 
HOI-Swap~\cite{hoi-swap} first replaces hand-interacting objects in a single frame based on one reference object image, and then employs a video diffusion model to generate the corresponding HOI video.
However, it overlooks the impact of object shape variations on the results. 
Re-HOLD~\cite{rehold} models HOI relationship through layout representations and introduces an adaptive layout adjustment strategy to handle shape variations, though its dependence on strict control signals limits its broader applicability.
In contrast, AnchorCraft~\cite{anchorcrafter} adopts multiple conditional inputs, including human poses, object depth, and hand meshes, to model HOI relationship, but its object-specific conditioning limits generalization.
HOMA~\cite{homa} explores HOI video generation under weak conditions, yet it lacks the capability for flexible HOI generation relying on body poses.

\noindent\textbf{Video Generation and Editing}.
With the rapid advancement of text-to-video models, many studies explore extending them for more controllable video generation and editing.
Early approaches relied on zero-shot~\cite{TokenFlow,Rerender,FateZero,DynVFX} or one-shot tuning learning paradigms~\cite{Tune-A-Vide, Shape-Aware, VideoSwap, Cut-and-Paste}, which limits both quality and generalization. 
Recent works aim to enhance model controllability through large-scale datasets and carefully designed architectures.
Some efforts~\cite{HuMo,phantom} leverage semantic information from reference images and inject it into the model to generate the corresponding subject video. 
VACE~\cite{vace} and HunyuanCustom~\cite{HunyuanCustom} unify diverse and complex multimodal inputs, enabling multiple video tasks such as subject swapping and human animation. 
However, these approaches struggle with complex hand–object interaction scenarios, often producing implausible objects and interactions.

\section{Methodology}
\label{sec:method}

\subsection{Preliminary and Formulation}

\noindent\textbf{Flow Matching DiT}.
Building upon recent advances in video generation~\cite{wanvideo,hunyuanvideo,yang2024cogvideox}, our work extends conditional DiT-based~\cite{peebles2023scalable} flow matching models~\cite{liu2022flow} to enable manipulable and controllable HOI synthesis. Given a 3D-VAE with encoder $\mathcal{E}$ and decoder $\mathcal{D}$, an input video 
$\textbf{V}$ 
is first mapped into a compressed latent representation $\textbf{Z} = \mathcal{E}(\textbf{V}) \in \mathbb{R}^{T\times C\times H\times W}$. 
The DiT-based model is then trained to estimate a time-dependent vector field that guides the noised latent video $\textbf{Z}_{t}$ toward the target data distribution $q(\textbf{Z}_{0}|c)$, where $c$ denotes the conditions (e.g., text prompt) and $t$ denotes the time step.
During inference, the DiT-based model evolves a noise latent sampled from a Gaussian distribution by predicting the velocity field $d\textbf{Z}_{t}/dt$, thereby transporting it toward the data manifold. 
The final video is then obtained by decoding the denoised latents via the VAE decoder, i.e., $\hat{\textbf{V}} = \mathcal{D}(\hat{\textbf{Z}}_{0})$.

\noindent\textbf{Task Definition}.
This work addresses the challenge of synthesizing versatile HOI scenarios by incorporating multiple forms of conditional controls.
Let the base text-to-video (T2V) model generate $\hat{\textbf{V}} = \operatorname{DiT}(\textbf{Z}_{t}, c_{\text{text}})$ given text condition $c_{\text{text}}$. We extend this formulation to
$\hat{\textbf{V}} = \operatorname{DiT}(\textbf{Z}_{t}, c_{\text{text}}, c_\text{visual}, c_\text{object}, c_\text{motion}, c_\text{bg})$, where $c_\text{visual}$, $c_\text{object}$, $c_\text{motion}$, and $c_\text{bg}$ represent visual appearance cues, object-specific information, motion constraints, and background conditions respectively. 
By flexibly activating and combining different instantiations of these conditioning signals, our framework supports a comprehensive range of input types, which are detailed in the subsequent sections.

\subsection{HOI Data and Model Input}
\label{sec:data_process}

The inherent scarcity of high-quality HOI video data represents a significant constraint on HOI synthesis. Therefore, we first provide a brief overview of our data curation and annotation process. For more detailed illustrations, please refer to the supplementary materials.

\noindent\textbf{Data Curation}.
The initial phase involves refining a large corpus of video clips to meet our specific quality and content requirements. We begin with a set of pre-segmented, single-shot videos and apply a three-stage filtering process.
\begin{itemize}
\item \textit{Score-Based Filtering}. Each video is evaluated by aesthetic, motion, and clarity scores, retaining only those surpassing all quality thresholds.

\item \textit{Human-Centric Filtering}. 
We retain human-centric, visually clear clips by filtering with human detection~\cite{ge2021yolox} and hand motion analysis~\cite{Jiang2023RTMPoseRM}, ensuring sharp, blur-free interactions.

\item \textit{VLM Filtering}. 
We use a VLM~\cite{Qwen2.5-VL} to filter videos where the subject holds a rigid object, yielding a clean dataset focused on rigid object interactions.

\end{itemize}

\noindent\textbf{HOI Data Process}.
After curating high-quality human-object interaction clips, we proceed with automated annotation to generate the final training data.

\begin{itemize}

\item \textit{Caption}. 
The VLM model~\cite{Qwen2.5-VL} is used to generate a concise yet descriptive text caption, which details the primary action, the person, and the object involved.

\item \textit{Wrist Trajectory}. 
From the full set of hand keypoints extracted during the human-centric filtering stage, we isolate the coordinates of the left and right wrists. 

\item \textit{Object Segmentation}. 
We then generate temporally consistent object masks using a multi-stage pipeline combining Grounding DINO~\cite{liu2024grounding}, pose-based disambiguation, SAM2~\cite{ravi2024sam} segmentation, and rigorous post-processing.
\end{itemize}

\noindent\textbf{Conditional Input}.
Based on the prepared HOI data, we detail four forms of conditioning that support our HOI synthesis objective.

\begin{figure*}[!t]
\centering
\includegraphics[width=\linewidth]{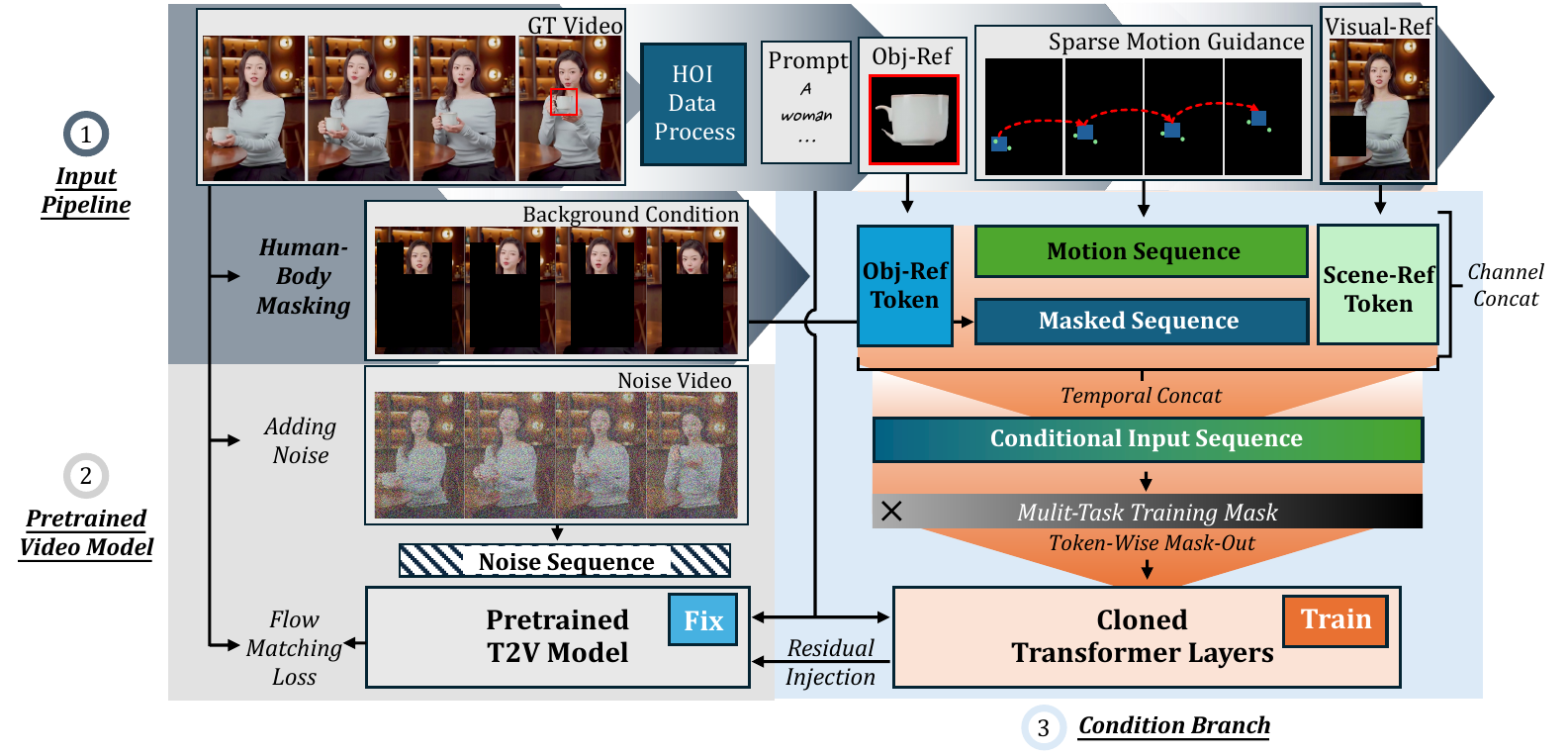}
\caption{
\textbf{The DISPLAY Framework}. 
We organize the proposed framework into three parts: \textit{Input Pipeline}, \textit{Pretrained Video Model}, and \textit{Condition Branch}.
1) In the \textit{Input Pipeline}, the input video is processed through a predefined HOI data procedure to extract the required multi-modal conditioning signals.
2) The \textit{Pretrained Video Model} preserves the original T2V denoising formulation.
3) The \textit{Condition Branch} encodes the multi-modal conditions and modulates the generation process via residual injection to guide the final video synthesis.
}
\label{fig:pipeline}
\end{figure*}

\begin{itemize}
    \item \textit{Object Reference}. 
    We randomly select one frame based on its visible area ratio \textit{w.r.t.} the object masks, favoring frames where the object is less occluded.
    
    \item \textit{Visual Reference}.
    We select the first frame of the input clip to provide global visual information, including both human appearance and background.
    Importantly, since training follows a self-reconstruction paradigm,
    we mask out the HOI region in the \textit{Visual Reference} to prevent unintended information leakage.
    
    \item \textit{Sparse Motion Guidance}.
    Although various forms of motion guidance have been explored in prior work~\cite{hoi-swap,anchorcrafter,rehold,homa} to model human–object interactions, they generally struggle to balance generation quality with flexibility. 
    Considering that objects may vary greatly in shape and scale, we instead use only the wrist points to guide hand trajectories. 
    This design is conceptually aligned with the ``end-effector''~\cite{tian2025emo2}, where the wrist moves tied primarily with overall human–object interactions.
    In addition, 
    we adopt a shape-agnostic yet size-aware bounding box to represent the object location. 
    
    \item \textit{Background Condition}.
    A single reference frame is sufficient to produce reasonable animation, yet it remains insufficient for performing localized edits in videos. 
    To address this, we optionally introduce a background frame sequence as an inpainting condition. 
    As illustrated in Fig.~\ref{fig:pipeline}, one configuration of this input masks out the human body.
    The choice of masking strategies is tied to our training strategy, which will be detailed in Sec.~\ref{sec:multi task training}.
    
\end{itemize}

\subsection{Architecture}
To effectively inherit the strong semantic understanding and generative capability acquired through massive-scale training of LVGM, we adopt a ControlNet~\cite{zhang2023adding}-style design, where the pretrained T2V model remains frozen and an additional conditioning branch is introduced for control injection. The overall architecture is depicted in Fig.~\ref{fig:pipeline}.

\noindent\textbf{Condition Branch}.
The Condition Branch is designed to handle multi-modal conditions, including 
text prompts and four kinds of conditional input introduced in Sec.~\ref{sec:data_process}.
Inspired by~\cite{vace,liang2025realismotion}, we introduce the \textit{Condition Branch} that clones a few transformer layers from the pretrained T2V backbone, applying only minimal modifications to preserve the original generative capacity while enabling effective conditioning.

Each \textit{Cloned Transformer Layer} is constructed by a cascaded self-cross attention design.
Accordingly, $c_{\text{text}}$ is incorporated following the standard pipeline via cross-attention. 
The \textit{Sparse Motion Guidance} is temporally aligned with the input video and encoded into the VAE latent space to obtain $c_{\text{motion}} \in \mathbb{R}^{N_{m}\times d_{m}}$ 
, where $N_{m}$ is the sequence length and $d_m$ is the channel dimension.
To enable localized video editing, we similarly encode a partially masked video (\textit{Background Condition}) into the latent space as $c_{\text{bg}}$.
We then concatenate $c_{\text{motion}}$ and $c_{\text{bg}}$ along the channel dimension, producing a feature that jointly provides temporal and spatial alignment constraints.

In addition, $c_\text{visual}$ and $c_\text{object}$ supply textural information for the overall scene and the target object, respectively. The \textit{Object Reference} and \textit{Visual Reference} are encoded using the same VAE encoder, and $c_\text{object}$ and $c_\text{visual}$ are concatenated with the previously obtained conditions along the sequential dimension.
The assembled input sequence for the \textit{Condition Branch} is represented as:
\begin{equation}
\begin{aligned}
\textbf{X}_c = \operatorname{S-Cat}[ & \operatorname{C-Cat}(c_\text{object},c_\text{object}); \operatorname{C-Cat}(c_{\text{motion}}, c_{\text{bg}}); \\
                      & \operatorname{C-Cat}(c_\text{visual}, c_\text{visual}) ],
\end{aligned}
\end{equation}
where $\operatorname{S-Cat}[\cdot;\cdot]$ and $\operatorname{C-Cat}(\cdot,\cdot)$ respectively represent sequential and channel concatenation.
Given a predefined set of layer indices $\mathbf{S}$, the overall condition is injected into the frozen video model as:
\begin{equation}
\begin{aligned}
\mathbf{Z}_{t}^{i'} = \operatorname{DiT}^{i}(\mathbf{Z}_{t},c_\text{text},t) &+  \operatorname{DiT}_{c}^{\operatorname{index}(\mathbf{S}, i)}(\mathbf{X}_{c},t), \\ 
i &\in \mathbf{S},
\end{aligned}
\end{equation}
where $\operatorname{DiT}^{i}$ is the $i$-th layer of the frozen video model and $\operatorname{DiT}_{c}^{\operatorname{index}(\mathbf{S}, i)}$ is the corresponding layer in the introduced \textit{Condition Branch}.

\noindent\textbf{Object-Stressed Attention}.
Unlike prior work~\cite{anchorcrafter,homa} that relies on visual understanding models such as DINO~\cite{oquab2023dinov2} and CLIP~\cite{radford2021learning}, we argue that using only VAE encodings of object textures—sharing the same latent space as the generated videos—is sufficient and enables more efficient learning for target recovery. 
Consequently, through sequential concatenation, object textures are progressively learned and integrated into the given video within the transformer layers via self-attention. 
The object boxes in the \textit{Sparse Motion Guidance} will provide spatial-temporal position and size priors, ensuring that the generated objects maintain appropriate appearance and motion consistency.

One remaining challenge is generating realistic interactions between the object and the hands given only the wrist positions provided in the \textit{Sparse Motion Guidance}. 
We empirically observe that emphasizing object-related attention improves object-hand interactions. Accordingly, we replace the standard self-attention operation with the Object-Stressed-Attention introduced below:
\begin{align}
& \text{Object-Stressed-Attention}([\textbf{x}_\text{obj}; \textbf{x}_\text{else}], \alpha) = \nonumber \\
& \operatorname{softmax}
\left (
\frac{1}{\sqrt{d}}
\begin{bmatrix}
\alpha^2\textbf{x}_\text{obj}\textbf{x}_\text{obj}^{\top}  & \alpha\textbf{x}_\text{obj}\textbf{x}_\text{else}^{\top} \\
\alpha\textbf{x}_\text{else}\textbf{x}_\text{obj}^{\top}  &\textbf{x}_\text{else}\textbf{x}_\text{else}^{\top}
\end{bmatrix}
\right )
[\textbf{x}_\text{obj}; \textbf{x}_\text{else}],
\end{align}
where $\alpha$ is a hyperparameter, $d$ is channel dimension, $\textbf{x}_\text{obj}$ is object tokens in the \textit{Conditional Input Sequence}, and $\textbf{x}_\text{else}$ denote others.

\noindent\textbf{Scene Consistency}.
Beyond object replacement in template videos, our goal is to generalize human-object interaction synthesis to animate and edit arbitrary reference frames or video templates. To this end, we introduce a visual reference frame capturing the entire scene, including human appearances and background. 
During training, this frame is encoded with the same VAE encoder and sequentially concatenated with other tokens, enabling the full visual information to be effectively integrated into the denoised results.

\subsection{Multi-Task Auxiliary Training}
\label{sec:multi task training}

Given the scarcity of high-quality HOI video data,
relying solely on high-quality HOI data would yield a severely limited training corpus. We empirically observe that this insufficiency causes degradation in generative results, specifically manifesting as reduced human structural fidelity and a deficit in overall interaction realism.
Nonetheless, video generation models can still substantially benefit from training on supplementary data lacking explicit HOI annotations, particularly for the synthesis of plausible human motion.
Therefore, we employ a multi-task training strategy that integrates both high-quality HOI-annotated data and videos with weak annotations.
This dynamic training is conducted through two configurations: the \textit{Human-Body Masking} strategy and the \textit{Multi-Task Training Mask}, as shown in Fig.~\ref{fig:pipeline}.

For the \textit{Human-Body Masking}, we consider two options: 
1) masking only the body region, as in Fig.~\ref{fig:pipeline}, and 
2) masking the entire frame. 
The first preserves the head position, stabilizing full-body motion, while the second offers greater generative flexibility.

For the \textit{Multi-Task Training Mask}, we apply a tensor sampled from a Bernoulli distribution to multiply the \textit{Motion Sequence} and \textit{Masked Sequence} before feeding them into the model. 
When parts of the \textit{Motion Sequence} are dropped, the model learns to synthesize plausible hand and object motions from the head–tail motion cues at both ends. The same principle applies to the \textit{Masked Sequence}.

In extreme cases where only the first-frame tokens are available, the model can be effectively trained in an image-to-video generation manner. 
Consequently, during inference, we can optionally omit the \textit{Background Condition} or \textit{Sparse Motion Guidance} to support image-to-video generation, video in-between synthesis, and HOI-related video editing.
More training details are discussed in the supplementary materials.

\section{Experiments}
\label{sec:experiment}

\noindent\textbf{Versatile Generation}.
To enable user-intended motion creation, we introduce a motion-authoring interface for generating the \textit{Sparse Motion Guidance} used during inference. 
Details of this interface are provided in the supplementary materials, where object and hand motions are specified using start and end positions along with their corresponding frame indices. Based on this design, our model supports three HOI generation scenarios:
\begin{itemize}
\item \textit{Object Replacement}.
When the target video contains an object, our method can directly replace it with a novel object without invoking the motion-authoring interface. 
\item \textit{Object Insertion}.
For scenarios requiring interaction with an object that does not exist in the original video, users can plan both the object placement and wrist trajectory via the introduced interface. 
\item \textit{Environmental Interaction}.
If an object is present in the video but not actively interacted with, users may employ our interface to define a desired hand-object interaction. 
\end{itemize}

\begin{table*}[t]
\centering
\caption{
\textbf{Quantitative Comparison.} 
Videos generated by our method and all baselines are evaluated from four perspectives, with appearance quality, temporal consistency, hand fidelity, and object fidelity assessed accordingly.
}
\resizebox{\textwidth}{!}{
\renewcommand{\arraystretch}{1.3}

\begin{tabular}{l|ccc|ccc|cc|cc}
\toprule
\multirow{2}{*}{Method} 
 & \multicolumn{3}{c|}{\textit{Appearance Quality}}       
 & \multicolumn{3}{c|}{\textit{Temporal Consistency}} 
 & \multicolumn{2}{c|}{\textit{Hand Fidelity}} 
 & \multicolumn{2}{c}{\textit{Object Fidelity}} 
 \\ 
\cline{2-11}
 & FID$\downarrow$  
 & LPIPS$\downarrow$ 
 & AES$\uparrow$ 
 & FVD$\downarrow$ 
 & MS$\uparrow$ 
 & SC$\uparrow$ 
 & HF$\uparrow$ 
 & CA$\uparrow$ 
 & O-CLIP$\uparrow$ 
 & O-DINO$\uparrow$ \\ 
\hline
VACE          
 & 105.708 & 0.134 & 0.524 
 & 862.65 & 0.995 & 0.953 
 & 0.972 & 0.871 & 0.802 & 0.677 \\
HunyuanCustom 
 & 72.185 & \textbf{0.060} & 0.541 
 & 742.34 & 0.994 & 0.956 
 & 0.975 & 0.832 & 0.770 & 0.603 \\
HuMo      
 & 254.620 & 0.552 & 0.464 
 & 1265.26 & 0.994 & \textbf{0.977}
 & 0.982 & - & 0.810 & 0.709 \\
Nano+WanAnimate   
 & 119.890 & 0.295 & 0.537 
 & 948.61 & 0.995 & 0.953 
 & \textbf{0.989} & 0.771 & 0.806 & 0.643 \\
Ours            
 & \textbf{67.501} & 0.072 & \textbf{0.547} 
 & \textbf{560.29} & \textbf{0.995} & 0.958
 & 0.987 & \textbf{0.891} & \textbf{0.890} & \textbf{0.832} \\
\bottomrule
\end{tabular}
}

\label{tab:cmp_sota_metrics}
\end{table*}

\begin{figure*}[t]
\centering
\includegraphics[width=\linewidth]{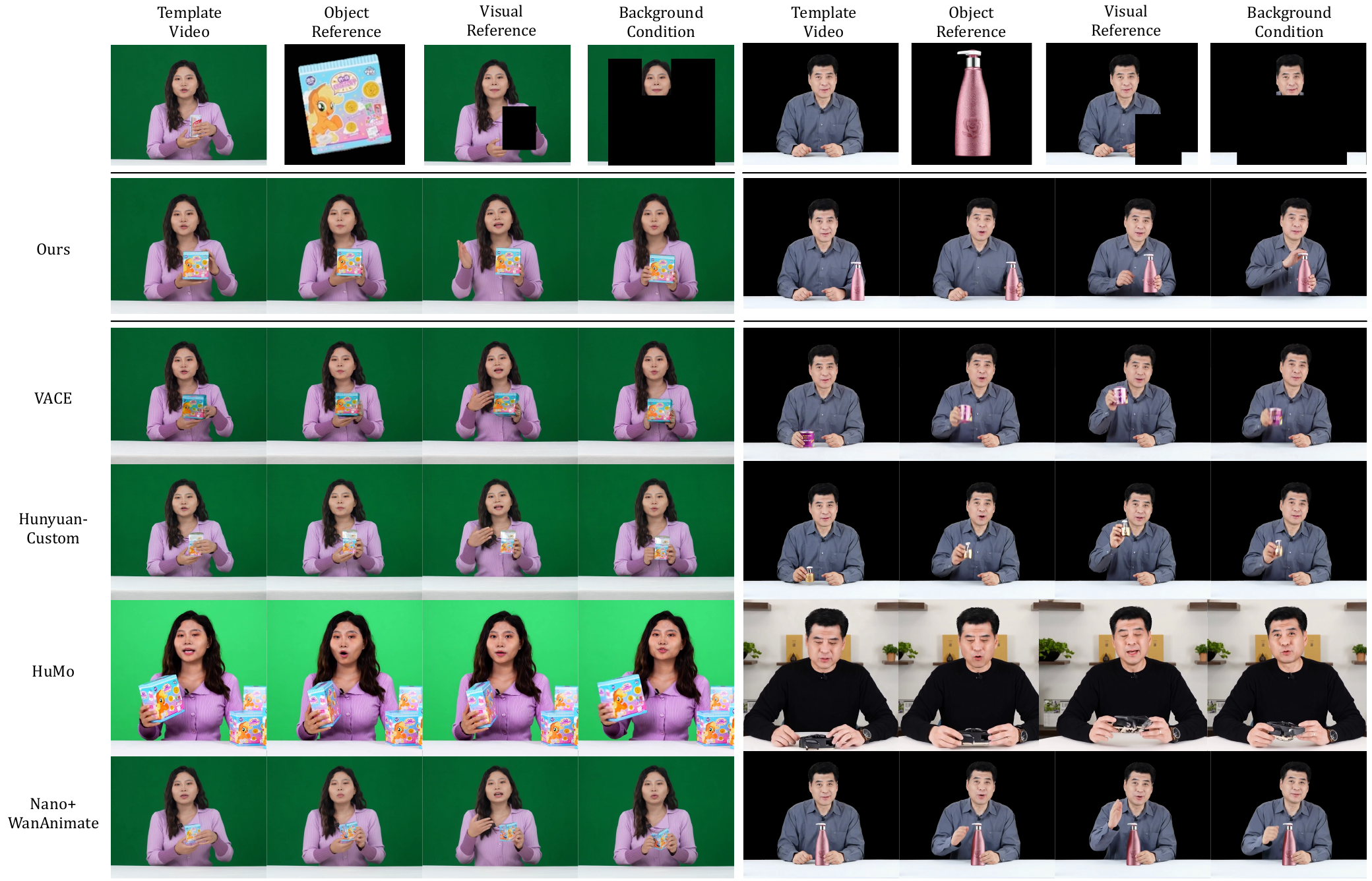}
\caption{
\textbf{Qualitative Comparisons}. 
We present an object-replacement comparison on the left, where the original object is substituted according to the provided object reference.
The visual reference and background condition used by our method are displayed in the top row.
On the right, we present an object-insertion comparison, where the template video contains no original object.
}
\label{fig:cmp_sota}
\end{figure*}

\begin{figure*}[t]
\centering
\includegraphics[width=0.75\linewidth]{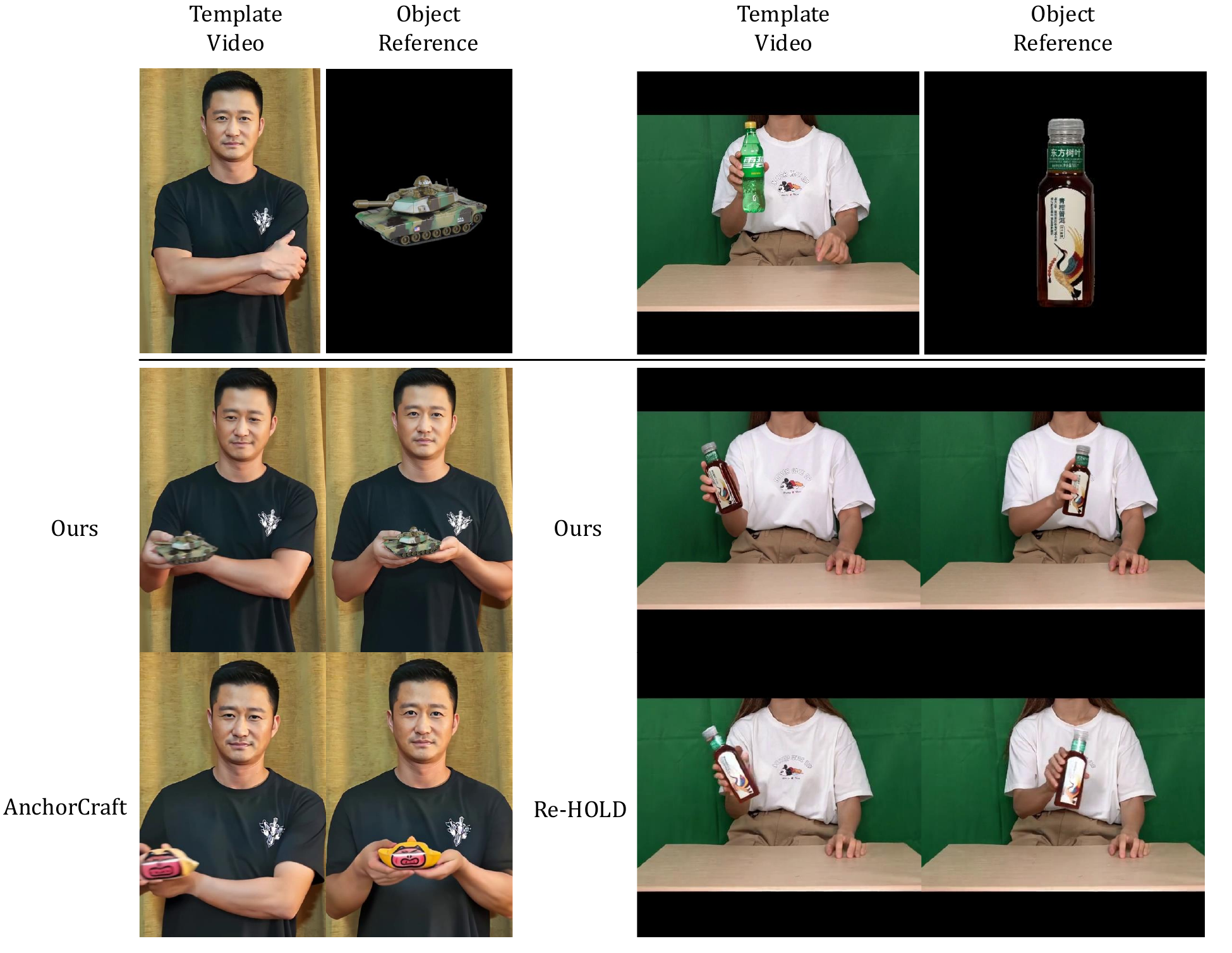}
\caption{\textbf{Qualitative Comparisons.}
Comparisons with AnchorCraft and Re-HOLD on their official results.}
\label{fig:rehold}
\end{figure*}

\noindent\textbf{Implementations.}
Our framework is built upon Wan2.1-14B~\cite{wanvideo}, which is used as a frozen T2V model. In parallel, we initialize evenly quarter-cloned transformer layers from Wan2.1 to form the trainable parameters of our \textit{Condition Branch}. 
The hyperparameter $\alpha$ of the introduced Object-Stressed Attention is set to 8. 
Our model is trained with a learning rate of $1e^{-5}$ with 32$\times$80G GPUs for two weeks.

\noindent\textbf{Datasets.}
We collect approximately 100 hours of HOI-annotated data using the proposed HOI data processing.
As detailed in Sec.~\ref{sec:multi task training}, an additional 50 hours of general human-centric videos are incorporated into multi-task training.
Evaluations are performed on our self-filmed test set and several in-the-wild video clips.

\noindent\textbf{Evaluation Metrics.}
We comprehensively evaluate the quality of results from four perspectives: 
1) \textit{Appearance Quality.} 
We evaluate the overall visual quality of the generated images by FID~\cite{fid}, LPIPS~\cite{lpips}, and Aesthetics (AES)~\cite{vbench}.
2) \textit{Temporal Consistency.} 
FVD~\cite{fvd}, Motion Smoothness (MS)~\cite{vbench}, and Subject Consistency (SC)~\cite{vbench} are employed to measure the temporal consistency of the videos.
3) \textit{Hand Fidelity} and
4) \textit{Object Fidelity}.
Following~\cite{hoi-swap}, we calculate Hand Fidelity (HF) and Contact Agreement (CA) to evaluate the quality of hand generation. 
In addition, we employ CLIP~\cite{radford2021learning} and DINO~\cite{oquab2023dinov2} models to compute the similarity between the reference and generated objects,
named Object-CLIP (O-CLIP) and Object-DINO (O-DINO).

\noindent\textbf{Comparison Methods.}
We compare our method with the following state-of-the-art approaches: 
VACE-14B~\cite{vace}, HunyuanCustom~\cite{HunyuanCustom}, HuMo~\cite{HuMo}, WanAnimate~\cite{Wan-Animate}, Re-HOLD~\cite{rehold}, and AnchorCraft~\cite{anchorcrafter}. 
Among these methods, VACE-14B and HunyuanCustom are unified models for video generation and editing.
Given a reference object image and a human video with a masked HOI region, both can generate videos depicting the person interacting with the reference object.
HuMo is a multi-subject video generation approach that produces HOI videos conditioned on the reference object and human images, along with text prompts.
WanAnimate focuses on human animation generation and does not natively support object replacement. 
To adapt to our task, we use image editing model Nano~\cite{nano} to replace the object in the first frame, and then use WanAnimate to generate the subsequent frames.
Since Re-HOLD and AnchorCraft do not support general HOI video generation, we conduct only comparisons with their official results.

\subsection{Quantitative Comparisons}
\label{sec:Quantitative Comparisons}
As shown in Table~\ref{tab:cmp_sota_metrics}, our method achieves the best FID and AES scores, along with the suboptimal LPIPS score, demonstrating its superiority in visual quality. 
Regarding temporal consistency, our method surpasses others in FVD, while achieving comparable SC and MS scores, reflecting strong temporal coherence and motion smoothness. 
Notably, HuMo attains the highest SC score due to its limited motion range in generated videos.
While the above metrics primarily evaluate overall generation quality, assessing the fidelity of hand and object synthesis is also crucial.
Benefiting from explicit hand motion supervision, WanAnimate achieves the best HF score, with our method performing comparably. 
In terms of the CA score, our method outperforms others, showcasing excellent HOI modeling capability. 
Moreover, the significantly higher O-CLIP and O-DINO scores demonstrate the strong object appearance preservation ability of our model, validating the effectiveness of the introduced Object-Stressed Attention.

\subsection{Qualitative Results}
\label{sec:Qualitative Results}

For qualitative comparison, we evaluate our method under two settings: object replacement and object insertion. 
As shown on the left of Fig.~\ref{fig:cmp_sota}, our method preserves the reference object's appearance with high fidelity while accurately modeling the hand–object interaction.
In contrast, other methods fail to faithfully preserve the reference object's texture and shape, often resulting in object deformation.
On the right of Fig.~\ref{fig:cmp_sota}, under the more challenging object insertion setting, methods like HuMo and HunyuanCustom struggle to generate realistic object textures, with HOI regions often showing noticeable artifacts.
While Nano+WanAnimate preserves object textures well, it remains constrained to the fixed motions given by the template video.
Benefiting from the introduced Object-Stressed Attention and Multi-Task Auxiliary Training strategies, our approach achieves satisfactory HOI generation, demonstrating superior realism and controllability.

Since Re-HOLD and AnchorCraft do not support general HOI video generation, we conduct comparisons with their official results in Fig.~\ref{fig:rehold}.
It can be seen that Anchorcraft, due to its overly strict control signals (i.e., hand meshes and object depth maps), overfits the textures of the training objects.
Re-HOLD suffers from limited model capacity, resulting in lower-quality HOI generation, particularly for the hands.
In contrast, our method produces satisfactory results.

\begin{figure*}[t]
\centering
\includegraphics[width=0.6\linewidth]{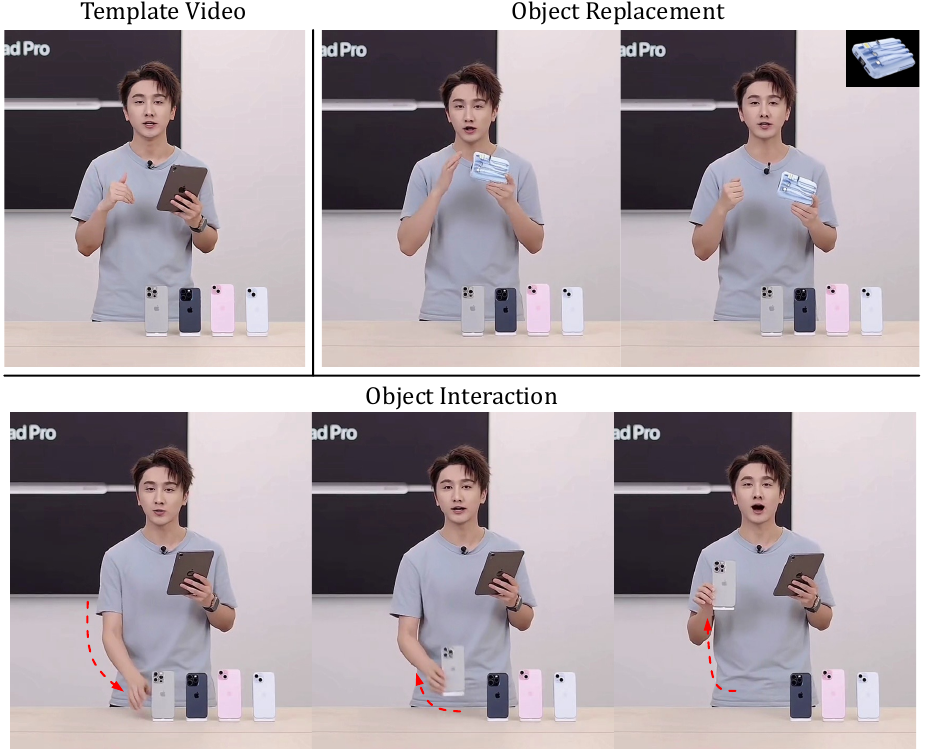}
\caption{ 
\textbf{Beyond Object Replacement}. Guided by sparse motions authored via the proposed user interface, our model supports environmental interactions. Red arrows indicate motion patterns we provide.}
\label{fig:multi}
\end{figure*}

\noindent\textbf{Beyond Object Replacement}.
Given our sparse-motion authoring scheme through a user-friendly interface, our method supports not only object replacement but also user-defined motion inputs for creating interactions with other objects in the scene.
As shown in Fig.~\ref{fig:multi}, our method can replace the iPad held by the person with a power bank, and also be able to guide the person to pick up the gray iPhone according to the specified motion conditions. 
This flexible HOI generation enables our approach to handle diverse real-world scenarios.

\begin{figure*}[t]
\centering
\includegraphics[width=\linewidth]{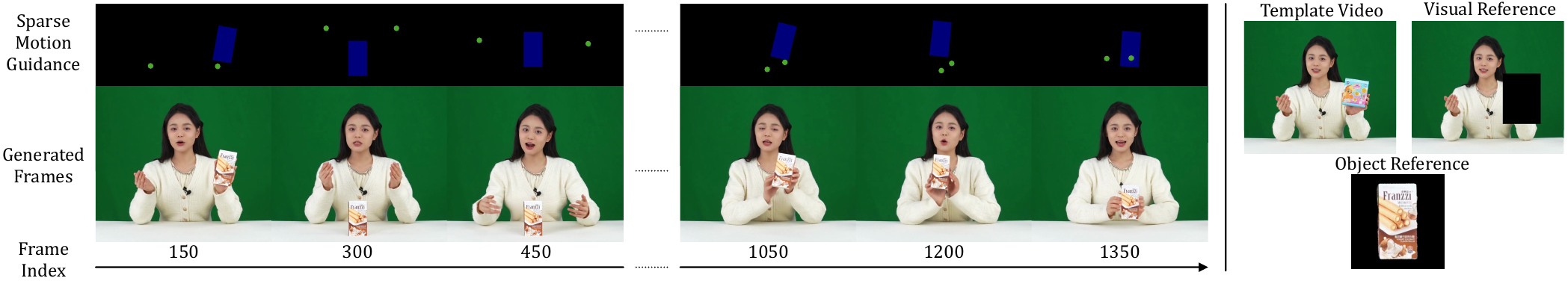}
\caption{
\textbf{Long-Video Manipulation}. 
Our model effectively generates long-video manipulations without noticeable error accumulation.
}
\label{fig:long}
\end{figure*}

\noindent\textbf{Long-Video Manipulation}.
By recursively feeding previously generated frames as conditions for subsequent clip generation, together with a consistent object reference, our method supports long-video manipulation.
Fig.~\ref{fig:long} presents a one-minute object-replacement result generated from a randomly chosen template video and object reference in our test set.
As shown, no noticeable error accumulation appears in the later frames.

\subsection{Ablation Study}
\label{sec:ablation}

\begin{table}[t]
\centering
\setlength{\tabcolsep}{6pt}
\renewcommand{\arraystretch}{1.15}
\caption{\textbf{Ablation study.} Quantitative results of different variants.}
\begin{tabular}{lcccc}
\toprule
Variation & FID $\downarrow$ & SC $\uparrow$ & HF $\uparrow$ & O-CLIP $\uparrow$ \\
\midrule
HSC        & 68.210 & 0.953 & 0.980 & 0.885 \\
w/o OSA    & 78.20  & 0.943 & 0.981 & 0.787 \\
w/o MTT    & 89.76  & 0.941 & 0.968 & 0.862 \\
w/o VR     & 107.64 & 0.947 & 0.985 & 0.865 \\
\textbf{Ours} & \textbf{67.501} & \textbf{0.958} & \textbf{0.987} & \textbf{0.890} \\
\bottomrule
\end{tabular}
\label{tab:ablation}
\end{table}

We evaluate our method against several variants to validate their effects, which are defined below. 
\begin{itemize}
\item  \textit{``Hand Skeleton Guidance (HSC)''}: We replace the wrist points with the hand skeleton.

\item \textit{ ``w/o Object-Stressed Attention (w/o OSA)''}: We replace Object-Stressed Attention with the vanilla self-attention operation.

\item \textit{``w/o Multi-Task Training (w/o MTT)''}: We do not adopt the multi-task training strategy.

\item \textit{``w/o Visual References (w/o VR)''}: Visual references are not employed as model inputs.
\end{itemize}

\begin{wrapfigure}{r}{0.5\textwidth}
\centering
\includegraphics[width=\linewidth]{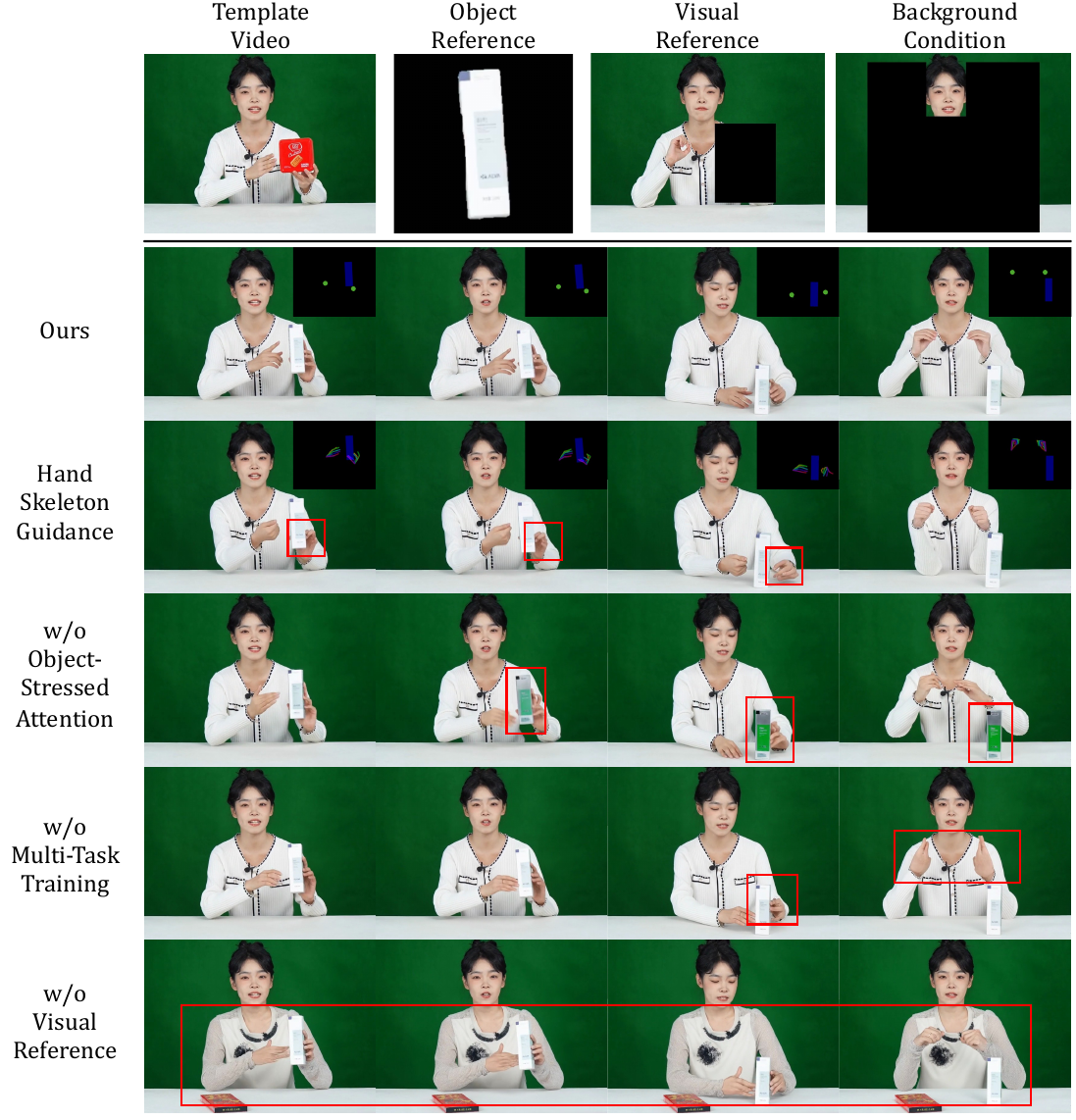}
\caption{ 
\textbf{Ablations.}
Qualitative results.}
\vspace{-10pt}
\label{fig:ablation}
\end{wrapfigure}

The quantitative and qualitative results are shown in Table~\ref{tab:ablation} and Fig.~\ref{fig:ablation}, respectively. It can be seen that using hand skeletons as motion guidance can produce the plausible hand movements, but struggles to adapt to the reference object when its shape differs from the original.
Replacing Object-Stressed Attention with self-attention severely affects the reference object's appearance quality and temporal consistency.
Without the multi-task training strategies on larger datasets, the quality of hand-region generation degrades.
Moreover, the absence of visual references provides insufficient information about human and scene appearance, resulting in poor appearance consistency.

\section{Conclusion}

In this paper, we present a novel framework, named DISPLAY, for controllable Human-Object Interaction (HOI) video generation.
By utilizing the \textit{Sparse Motion Guidance}, our method achieves a balanced representation of human and object dynamics, effectively enabling robust synthesis with novel objects.
Especially, the \textit{Object-Stressed Attention} mechanism reinforces spatial coherence and physical plausibility.
Moreover, the \textit{Multi-Task Auxiliary Training} strategy further improves generalization when high-quality HOI data is limited.
Our framework supports diverse applications such as object insertion, replacement, and environment-aware interaction, offering an intuitive and flexible paradigm for HOI generation.

\clearpage

\bibliographystyle{splncs04}
\bibliography{main}

\appendix
\section*{Appendix}
\section{Training Details}

As described in the \textit{Multi-Task Auxiliary Training} section, our model is trained on both HOI-annotated data and human videos with weak annotations.
Based on whether the held object is annotated, the training corpus is divided into two categories: HOI-annotated data and human videos with weak annotations (wrist points).
When training samples lack object annotations or contain no human–object interaction, we set the object reference token to an all-zero placeholder and omit the object bounding box from the \textit{Sparse Motion Guidance}.

Moreover, for the \textit{Human-Body Masking} strategy, we adopt two options:
1) masking only the body region, and
2) masking the entire frame.
The first option preserves the head position, helping stabilize full-body motion, while the second provides greater generative freedom. Accordingly, during inference, the first option supports human-centric video editing, enabling the model to reconstruct masked regions from any template video, while the second option empowers the model to generate full-frame videos.

For the \textit{Multi-Task Training Mask}, we sample a Bernoulli tensor to multiply the \textit{Motion Sequence} and \textit{Masked Sequence} before feeding them into the model.
This design offers several advantages:
1) Dropping the \textit{Masked Sequence} except for the first frame corresponds to a single-frame animation or image-to-video setting, depending on whether the \textit{Motion Sequence} is provided. 
Thus, at inference time, our model supports video generation from a single reference frame and video inpainting given the head and tail clips.
2) Since the \textit{Motion Sequence} is authored via the proposed interface, its basic form may limit the naturalness of human motion. During inference, we can therefore drop intermediate segments while retaining the start and end positions. Reducing reliance on this strong condition also enhances text-prompt controllability and increases generative flexibility.

\begin{figure*}[]
\centering
\includegraphics[width=.8\linewidth]{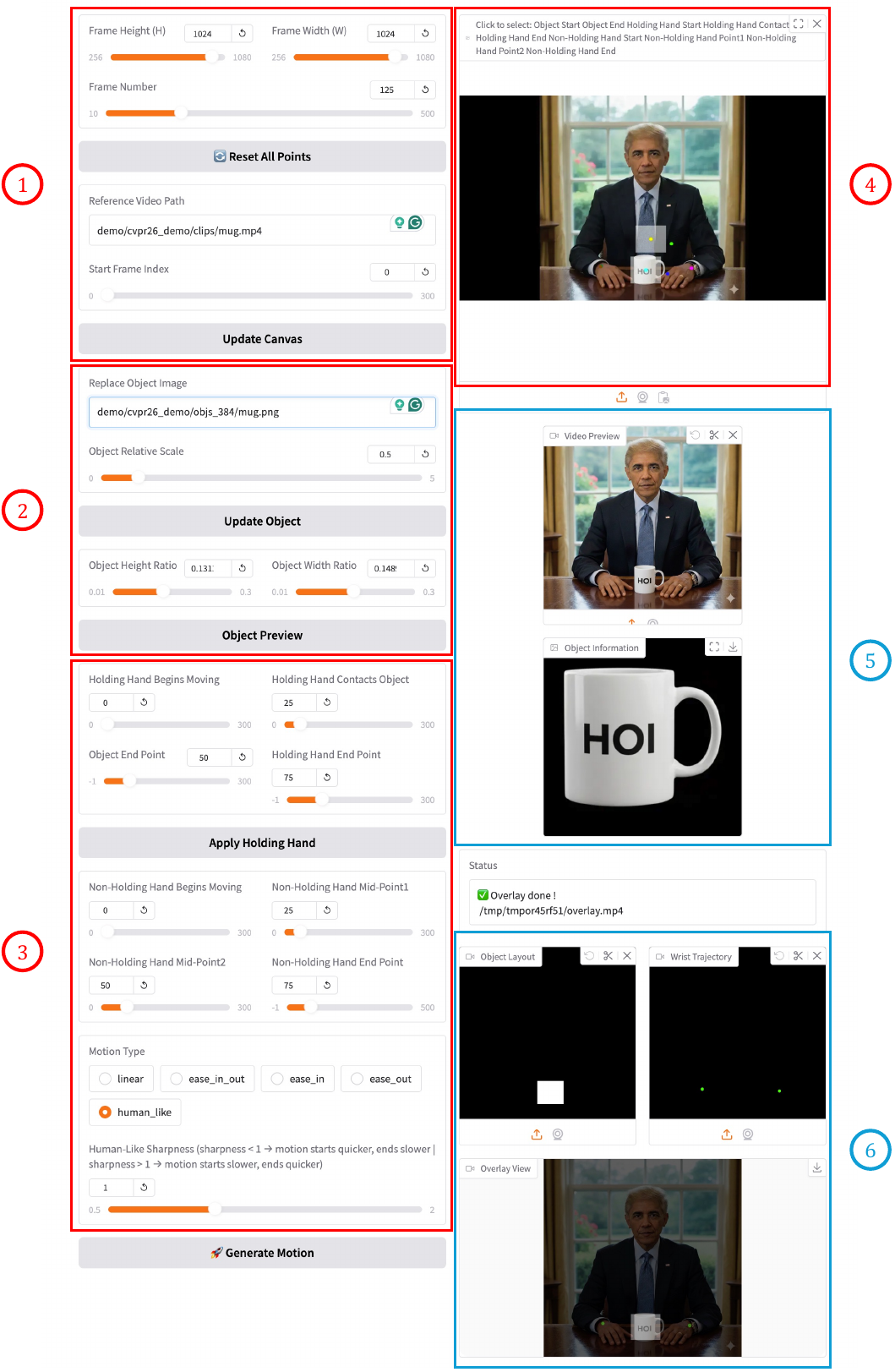}
\caption{
\textbf{Motion-Authoring Interface}. 
We present the proposed user interface, which facilitates the creation of \textit{Sparse Motion Guidance} for high-freedom HOI generation.
In general, the red boxes display the input components, while the blue boxes show the input references and the authored motions.
}
\label{fig:interface}
\end{figure*}

\section{Motion-Authoring Interface}
To enable user-intended motion creation, we introduce a motion-authoring interface for generating the \textit{Sparse Motion Guidance} used during inference. 
Fig.~\ref{fig:interface} presents the overall layout of the interface.
Here, we briefly describe each component alongside three HOI generation scenarios, and provide a dynamic demonstration in the supplementary video.

As shown in Fig.~\ref{fig:interface}, the interface is divided into six components, with red elements indicating inputs and blue elements indicating outputs.
The \textcolor{red}{\textbf{\textcircled{1}}} section provides basic information about the template video/image, including frame width, height, and the number of frames to generate, along with a slider to select the start frame for video inputs.
The \textcolor{red}{\textbf{\textcircled{2}}} section specifies the object input. A slider adjusts the object size or sets it as a relative ratio to the frame dimensions, ensuring proper placement for objects of various shapes.
The \textcolor{red}{\textbf{\textcircled{3}}} and \textcolor{red}{\textbf{\textcircled{4}}} sections provide a simple scheme for trajectory creation. Users specify the start, midpoint, and end positions via clicks on the canvas in \textcolor{red}{\textbf{\textcircled{4}}}, and assign the corresponding frame indices in \textcolor{red}{\textbf{\textcircled{3}}}, ensuring coherent motion paths and timing.
The \textcolor{blue}{\textbf{\textcircled{5}}} section provides an overview of the input video/image template, while the \textcolor{blue}{\textbf{\textcircled{6}}} section displays three output previews: the wrist trajectory, the bounding-box trajectory, and their combined overlay.
Based on this design, our model supports three HOI generation scenarios:

\begin{itemize}
\item \textit{Object Replacement}.
When the target video contains an object, our method can directly replace it with a novel object without invoking the motion-authoring interface. 
A scale adjustment ensures that the substituted object fits naturally into the scene \textit{w.r.t.} the object size. 
Notably, even when the new object differs significantly in shape from the original, our wrist-based motion guidance enables the hands to adapt seamlessly, requiring no additional shape-specific tuning.
\item \textit{Object Insertion}.
For scenarios requiring interaction with an object that does not exist in the original video, users can plan both the object placement and wrist trajectory via the introduced interface. 
This enables the synthesis of natural hand-object interactions, such as picking up the object, manipulating it for display, and placing it back down.
\item \textit{Environmental Interaction}.
If an object is present in the video but not actively interacted with, users may employ our interface to define a desired hand-object interaction. 
The interface allows selection of the target object and specification of its spatial layout and wrist trajectories. These trajectories are defined simply using start and end positions together with corresponding frame indices.
\end{itemize}

\begin{figure}[t]
\centering
\includegraphics[width=0.7\linewidth]{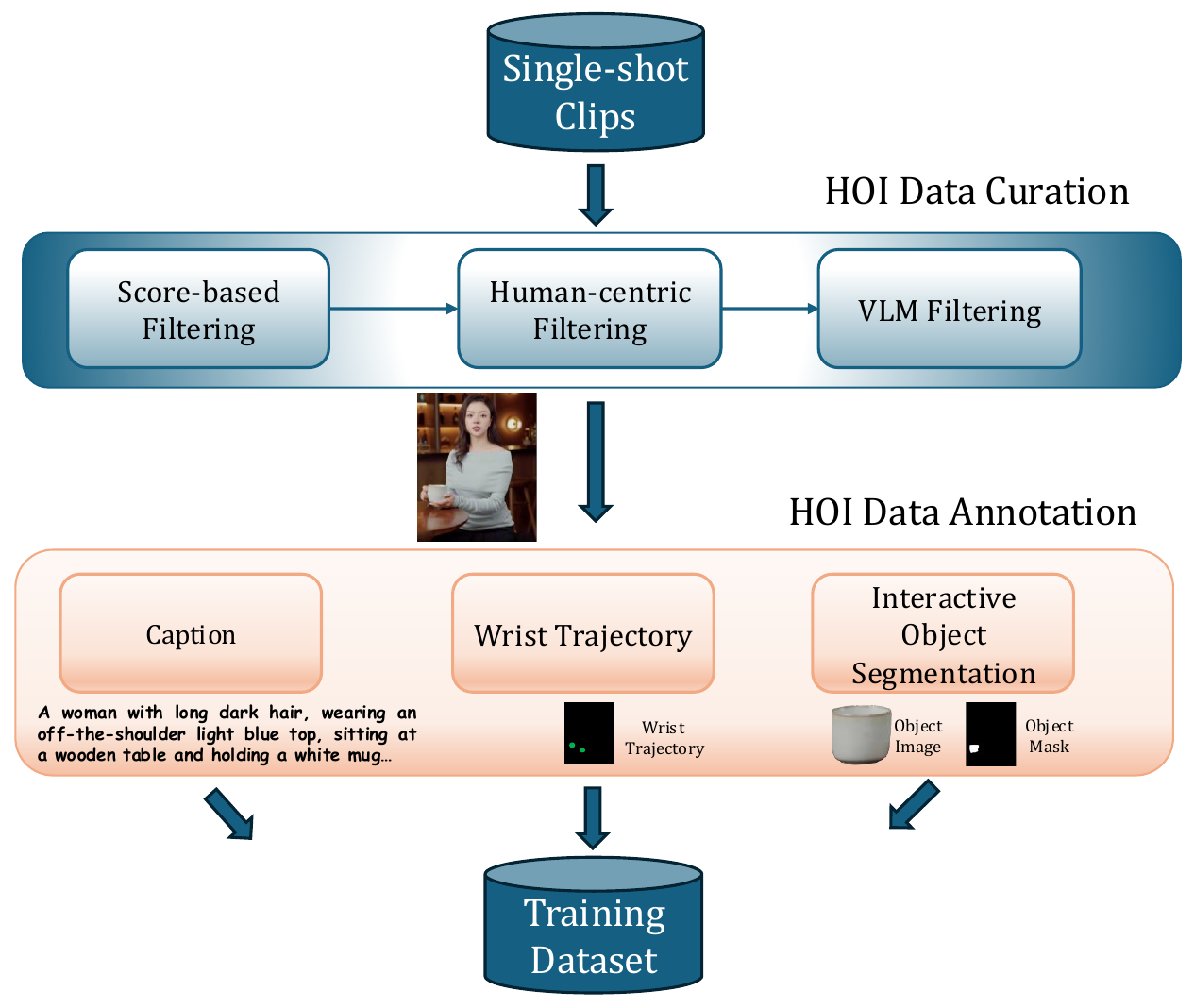}
\caption{ 
\textbf{HOI data construction pipeline.}
Through HOI Data Curation and Annotation, high-quality HOI data is obtained for model training.}
\label{fig:data}
\end{figure}

\section{HOI Data Construction}
A high-quality Human-Object Interaction (HOI) dataset is crucial for ensuring model performance. Therefore, we meticulously designed a comprehensive HOI dataset construction pipeline.
Our initial video data is sourced from in-the-wild web videos and self-collected green screen footage.
We begin by transcoding all video data to 25 frames per second (fps).
Next, we perform scene detection and segmentation to ensure each resulting video segment is a single-shot scene.
Finally, the filtered videos are segmented into 5-second clips to form the initial pool of raw video fragments.

Based on these steps, we amassed a large collection of video clips. 
Our subsequent process primarily involves two stages, as illustrated in Fig.~\ref{fig:data}:
1) \textit{HOI Data Curation}: The objective of this stage is to filter out high-quality HOI samples from the vast quantity of raw video clips.
2) \textit{HOI Data Annotation}: This stage is dedicated to annotating the HOI data, which ultimately provides high-quality training data for subsequent model training.
Next, we will introduce these two stages in detail.
\subsection{HOI Data Curation}
\begin{itemize}
    \item \textit{Score-Based Filtering}. 
    Low-quality data present in the raw clips (e.g., blurry videos or videos with excessively slow or fast motion) can negatively impact model performance. Therefore, we filter the video clips using three metrics: Aesthetic Score~\cite{vbench}, Motion Score~\cite{motionscore}, and Clarity Score, retaining only those clips whose scores exceed a predefined threshold.
    \item \textit{Human-Centric Filtering}. 
    As our task focuses on human-centric generation, we must address the presence of video clips in the database that are irrelevant to human content. 
    To this end, we employ human detection~\cite{ge2021yolox}  and hand motion analysis~\cite{Jiang2023RTMPoseRM} to filter the data, ensuring the remaining videos feature clear and fluid human motion.
    \item \textit{VLM Filtering}.
    After obtaining the human-centric videos, the subsequent step is to select suitable Hand-Object Interaction (HOI) videos. 
    We utilize a Vision-Language Model (VLM)~\cite{Qwen2.5-VL}, guided by appropriate text prompts, to filter for videos where a person is holding a rigid object. 
    Following the aforementioned steps, we obtain a clean, high-quality HOI dataset.
\end{itemize}

\subsection{HOI Data Annotation}
\begin{itemize}
    \item \textit{Caption}.
    Our method is built upon a text-to-video foundation model, where a text prompt is required.
    We thus use a VLM~\cite{Qwen2.5-VL} to comprehend the video content and generate a concise yet descriptive text, covering the background, person/object appearance, and the human-object interaction motion.
    \item \textit{Wrist Trajectory}.
    From the full set of hand keypoints obtained during the human-centric filtering stage, we isolate the coordinates of the left and right wrists. 
    These coordinates are then combined with the subsequently obtained object masks to derive sparse motion guidance.
    \item \textit{Object Segmentation}.
    We need to accurately segment the object interacting with the hand, rather than any unrelated objects.
    To achieve this, we first obtain all object bounding boxes in the scene using Grounding DINO~\cite{liu2024grounding}. 
    After filtering out objects with unreasonable sizes, we select the one closest to the hand keypoints as the interaction object.
    We then apply SAM2~\cite{ravi2024sam} to segment this interaction object and obtain its corresponding object mask and object video.
    It is worth noting that we sort the object video frames based on the size of the mask areas.

\end{itemize}

\section{Experimental Settings}
Our comparison methods include VACE-14B~\cite{vace}, HunyuanCustom~\cite{HunyuanCustom}, HuMo~\cite{HuMo}, WanAnimate~\cite{Wan-Animate}, Re-HOLD~\cite{rehold}, and AnchorCraft~\cite{anchorcrafter}.
For VACE-14B and HunyuanCustom, we use their \textit{Subject Reference + Inpainting} functionality to perform HOI generation.
Specifically, we first mask out the original object region in the template video using the corresponding object mask, then provide the masked video together with the reference object image as inputs to generate the HOI video with the replaced object.
For HuMo, we provide a clean human image without any objects, along with the reference object image, an appropriate text prompt, and audio as inputs to generate the HOI video.
The goal is for the person and the object to retain their original appearances while following the text prompt to perform plausible hand–object interaction movements.
For Wananimate, we first feed the first frame of the template video together with the reference object image into the image editing model Nano~\cite{nano} and provide an appropriate prompt to replace the original object in the frame with the reference object.
The edited image is then used as the reference image, while the skeleton sequence of the template video is used as the motion guidance.
Both are fed together into Wananimate to generate the HOI video.

\section{Supplementary Video}
To provide a more intuitive understanding of our method and its qualitative comparisons with SOTAs, we include a video demo in the supplementary materials.
The demo is divided into eight sections, as outlined below:
\begin{itemize}
    \item \textit{Introduction}.
    A showcase of how our method leverages an image generation model~\cite{nano} to create imaginative human–object interaction videos.
    \item \textit{Object Replacement Results}.
    Demonstrates the results of our method and SOTAs under the Object Replacement setting.
    \item \textit{Object Insertion Results}.
     Presents the results of our method and SOTAs under the Object Insertion setting.
    \item \textit{Environmental Interaction Results}.
    Highlights our method’s ability to generate interactions with arbitrary objects in the video.
    \item \textit{Ablation Experiments}.
    Compares the HOI generation quality of our method under different ablated variants.
    \item \textit{Other Comparisons}.
    Shows our results alongside the official results of AnchorCraft and Re-Hold.
    \item \textit{Long Video Manipulation}.
    Demonstrates our method’s performance on long video generation.
    \item \textit{Motion-Authoring Interface Introduction}.
    Introduces the workflow of the proposed Motion-Authoring Interface.
\end{itemize}

\section{Limitations}
While our framework achieves controllable and high-quality HOI synthesis, it still faces some limitations. Non-rigid objects: The proposed sparse guidance does not model deformation. Thus, interactions involving soft or highly deformable objects may exhibit unrealistic shapes or motion.
Complex object geometry: Since our training data relies on SAM-based segmentation, object masks for geometrically complex or deformable objects are often incomplete. 
As a result, our framework struggles to accurately preserve detailed object shapes under these conditions.
These limitations point to future directions such as incorporating deformation-aware guidance or stronger object representations.

\section{Broader Impact}
Our method focuses on generating highly realistic human–object interaction videos, significantly advancing and enriching the applications of digital human technologies in areas such as e-commerce livestreaming and entertainment.
However, hyper-realistic generated videos may also have negative societal impacts if misused, such as in telecom fraud or the spread of disinformation.
Therefore, we will strictly monitor the use and dissemination of our model and its generated content to ensure it is employed only for academic research or positive purposes.
    
\end{document}